\documentclass[letterpaper, 10 pt, conference]{ieeeconf}  % Comment this line out if you need a4paper
\IEEEoverridecommandlockouts                              % This command is only needed if you want to use the \thanks command
\IEEEoverridecommandlockouts
\usepackage{cite}
\usepackage{amsmath,amssymb,amsfonts}
\usepackage{graphicx}
\usepackage{textcomp}
\usepackage[ruled,linesnumbered]{algorithm2e}
\usepackage{amsmath}
\usepackage{xcolor}
\usepackage{hyperref}
\usepackage{booktabs}
\usepackage[T1]{fontenc}
\usepackage{aecompl}

\SetCommentSty{mycommfont}

\definecolor{ao(english)}{rgb}{0.0, 0.5, 0.0}
\hypersetup{
    colorlinks=true,
    linkcolor=green,
    filecolor=magenta,      
    urlcolor=ao(english),
}

\def\BibTeX{{\rm B\kern-.05em{\sc i\kern-.025em b}\kern-.08em
		T\kern-.1667em\lower.7ex\hbox{E}\kern-.125emX}}
\begin{document}
	
	\title{A General Framework for Lifelong Localization and Mapping in Changing Environment}
	\author{Min Zhao$^{1}$, Xin Guo$^{1}$, Le Song$^{1}$, Baoxing Qin$^{1}$, Xuesong Shi$^{2}$, Gim Hee Lee$^{3}$, Guanghui Sun$^{4}$%
		\thanks{*This work was supported in part by the National Key R\&D Program of China (No. 2019YFB1312000).}% <-this % stops a space
		\thanks{$^{1}$Gaussian Robotics, Shanghai, China.}
		\thanks{$^{2}$Intel Labs China, Beijing, China.}%
		\thanks{$^{3}$Computer Vision and Robotic Perception Lab, Department of Computer Science, School of Computing, National University of Singapore, Singapore.}%
		\thanks{$^{4}$Department of Control Science and Engineering, Harbin Institute of Technology, China.}%
		\thanks{$^{1}$Author to whom correspondence should be addressed, E-mail,
			{\tt\small zhaomin@gs-robot.com}}}
	\maketitle
	\begin{abstract}
		The environment of most real-world scenarios such as malls and supermarkets changes at all times. A pre-built map that does not account for these changes becomes out-of-date easily. Therefore, it is necessary to have an up-to-date model of the environment to facilitate long-term operation of a robot. To this end, this paper presents a general lifelong simultaneous localization and mapping (SLAM) framework. Our framework uses a multiple session map representation, and exploits an efficient map updating strategy that includes map building, pose graph refinement and sparsification. To mitigate the unbounded increase of memory usage, we propose a map-trimming method based on the Chow-Liu maximum-mutual-information spanning tree. The proposed SLAM framework has been comprehensively validated by over a month of robot deployment in real supermarket environment. Furthermore, we release the dataset collected from the indoor and outdoor changing environment with the hope to accelerate lifelong SLAM research in the community. Our dataset is available at \url{https://github.com/sanduan168/lifelong-SLAM-dataset}.
	\end{abstract}
	
	\section{Introduction}
\begin{figure}[htpb]
		\centering
		\includegraphics[scale=0.44]{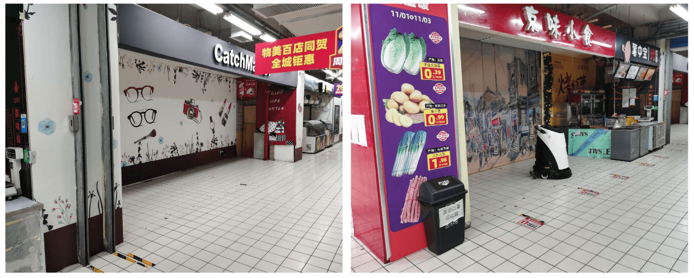}
		\caption{The images above captured from the same perspective at different time.}
		\label{fig:environment}
	\end{figure}

	Accurate and robust localization is one fundamental requirement for robot navigation in complex environments. Fig. \ref{fig:environment} shows a service robot working in a dynamic environment of a supermarket. The scenes are full of non-static objects, e.g. doors, goods and containers, which can move anytime. With a static map, conventional scan matching methods \cite{b1}, \cite{b2}, \cite{b3}, \cite{b4} may fail easily when a robot traverses through a changing environment. Moreover, the wrong matching result may introduce erroneous loop closure measurements and further leads to catastrophic failures in the back-end optimization \cite{b5}. To handle such problems, the robot needs to have the ability to construct and maintain an up-to-date map while it continuously operates in a changing environment. We refer to this as the lifelong Simultaneous Localization and Mapping (SLAM) problem. 
	
	A typical SLAM system consists of a front-end and a back-end module \cite{b5}, \cite{b6}, \cite{b7}. The front-end module collects data such as LIDAR point clouds and camera images from sensors and finds the transformation between consecutive data frames. The back-end module corrects the front-end estimation drifts by performing the loop closures. 		
	In order to tackle environment changes, we introduce a map updating module on top of the front-end and back-end modules. This map updating module performs the following:
	
	\begin{itemize}
		\item Collect sensor data and record the dynamic scenes while doing localization.
		\item Detect the differences between old and real-time updated maps.
		\item Trim the old maps with real-time updated maps, and thus keeping pace with the environment changes under an efficient constant computation complexity.
	\end{itemize}
	
	\begin{figure*}[]
		\centering
		\includegraphics[scale=0.2]{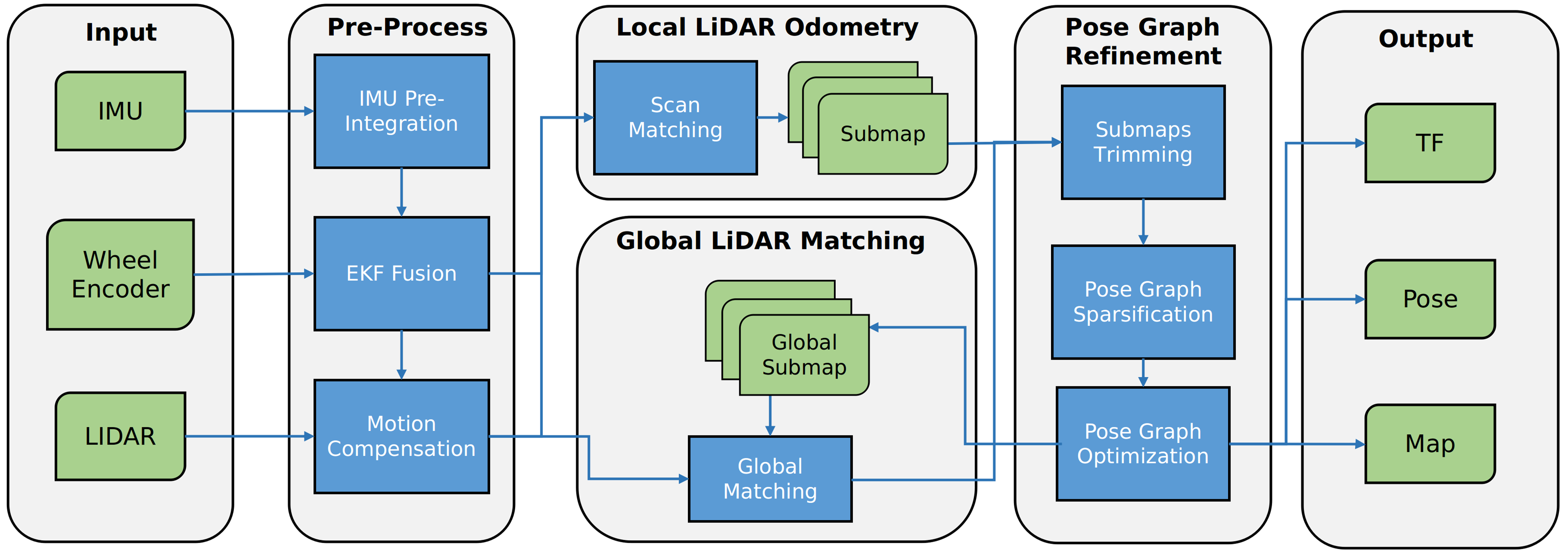}
		\caption{An overview of the proposed lifelong SLAM framework}
		\label{fig:framework}
	\end{figure*}
	
	In this paper, we propose a general framework for lifelong localization and mapping. Specifically, our framework tracks the changes in the scene and maintains an up-to-date map for accurate and robust localization estimation.	We test our method on real commercial robots that operate in a supermarket environment continuously for more than a month. The experiment results show that our method can achieve accurate and robust localization in the presence of significant environment changes.
	
	Our main contributions are summarized as follow:
	
	\begin{itemize}
		\item A complete general framework for lifelong SLAM, which operates effectively in changing environment.
		\item A submap-based graph sparsification method that achieves high accuracy with constant computation and memory complexity.
		\item A public dataset of LIDAR, IMU and wheel encoder data in changing environments for lifelong SLAM.
		
	\end{itemize}
	
	\section{Related Works}
	
	\subsection{Lifelong SLAM}\label{A}
	
	Long-running robots need to consider mapping the changing environment as a life-long process:
	deleting old nodes and adding new features to update the map\cite{b8}, \cite{b9}, \cite{b10} and \cite{b11}. Biber et al. \cite{b8} presents a kind of  dynamic map. The key technical contribution is the use of a sample-based representation and its interpretation through robust statistics. The short-term memory map is updated after each localization step and the long-term map is stored and evaluated only after a run or after a day. Walcott et al. \cite{b10} presents Dynamic Pose Graph SLAM (DPG-SLAM), an algorithm designed to remove the inactive scans and add the new scans to the map by detecting the label changes after each pass. However, they maintain a pose graph which is comprised of all the scan nodes that will not be optimized in real-time because of high computation complexity. Ding et al. \cite{b11} proposes a joint framework for vehicle localization that adaptively incorporate results from both the LiDAR inertial odometry and global matching modules. To overcome the failure of global matching, they also propose an environmental change detection method to find out when and what portion of the map should be promptly updated. 

	In contrast to them, for the purpose of decreasing computational complexity, our method do not include the map change detection module. The map data structure of ours is comprised of several submaps and we replace the old submaps with new ones directly.
	
	\subsection{Pose Graph Sparsification}
	Another task for lifelong SLAM is to sparse the pose graph while minimizing the loss of information when the robots revisit already mapped area. The computation complexity and memory usage should be proportional to the scale of explored space rather than to the localization duration. To this end, pose graph sparsification methods are proposed. Kretzschmar et al. \cite{b12} uses the expected information gain, which is defined as the expected reduction of uncertainty in the belief of the robot caused by an observation, to reduce the redundant nodes and keep constant size pose graph. In \cite{b13}, the authors propose the use of a Chow-Liu tree (CLT) \cite{b14} to approximate the individual elimination cliques as sparse tree structures and seek to minimize the loss of information, restricting the size of the pose graph. Carlevaris et al. \cite{b14} introduces a generic factor-based method for node removal in factor-graph SLAM, which is referred to as generic linear constraint(GLC). It operates on the Markov blanket of a marginalized node and compute a sparse approximation of the blanket. In this work, we combine the CLT sparsification method with submap trimming and  resulting in the lifelong SLAM pipeline as below.
	
	\section{System Overview}
	\subsection{System Structure}
	This section describes the architecture of our proposed framework as shown in Fig. \ref{fig:framework}. Our system consists of three subsystems: local LiDAR odometry (LLO), global LiDAR matching (GLM) and pose graph refinement (PGR). The role of LLO is the same with \cite{b2}: building a succession of submaps which are locally consistent. The GLM subsystem is responsible for calculating the relative constraint between the incoming scans and global submaps, as well as inserting the submaps and constraints into PGR. PGR is the most important part of our system. It collects the submaps from LLO and constraints from GLM, trims the old submaps saved in old session, and executes pose graph sparsification and optimization. These modules will be discussed in detail in Section \uppercase\expandafter{\romannumeral4}. 
	
	\subsection{Notation}
	In this paper, we use $\mathcal{X}$, $\mathcal{M}$, $\mathcal{Z}$ to represent the node poses, submap poses and relative constraints, respectively. Due to multi-session localization, each session consists of several nodes, submaps and constraints. $\mathcal{X}$ is the collection of $\mathbf{x}_{i}^{s}$ , which $i$ is the index of the node pose and $s$ means the session ID. Analogously, $\mathcal{M}$ is composed by $\mathbf{m}_{i}^{s}$. It should be emphasized that there are three types of constraints in our system: node-to-submap $\mathbf{z}_{ij}^{mix}$, node-to-node $\mathbf{z}_{ij}^{node}$ and submap-to-submap $\mathbf{z}_{ij}^{map}$. %These constraints are denoted as $\mathbf{z}_{ij}^{mix}$, $\mathbf{z}_{ij}^{node}$ and $\mathbf{z}_{ij}^{map}$, respectively. 
	These constraints take the form of relative poses and associated covariance matrice $\mathbf{\Omega}_{ij}^{mix}$, $\mathbf{\Omega}_{ij}^{node}$ and $\mathbf{\Omega}_{ij}^{map}$. For example, $\mathbf{z}_{i^{s}j^{s'}}^{mix}$ is the relative transform between the node pose $\mathbf{x}_{i}^{s}$ and submap pose $\mathbf{m}_{j}^{s'}$. The adjacent two node poses are related by the odometry constraint, e.g. $\mathbf{z}_{i^{s}j^{s'}}^{odom}$ controls $\mathbf{x}_{i}^{s}$ and $\mathbf{x}_{j}^{s'}$. Specifically, the pose and constraint is represented by a rotation $\mathbf{R} \in \text{SO}(2)$ and a translation $\mathbf{t} \in R^{2}$ because of the flat terrain of indoor environment. Nonetheless, our framework 
	can be easily extended to $\text{SE}(3)$.
	
	\subsection{Problem Definition}
	Given the poses and relative measurement constraints, the pose graph optimization is formulated as a maximum a posteriori probability (MAP) estimation problem. We then convert the MAP problem into a least square problem with the following cost function:
	\begin{equation}
	%	\begin{aligned}
	\mathcal{X}^{*},\mathcal{M}^{*} = \operatorname*{argmin}_{\mathcal{X},\mathcal{M}} J(\mathcal{X},\mathcal{M},\mathcal{Z}).
	%	\end{aligned}
	\end{equation}
	Note that $\|\mathbf{x}\|_{\mathbf{\Omega}}^{2}=\mathbf{x}^{\top }\mathbf{\Omega}^{-1}\mathbf{x}$, and $J(\mathcal{X},\mathcal{M},\mathcal{Z})$ is defined by the following equation:
	\begin{equation}
	\begin{split}
	J(\mathcal{X},\mathcal{M},\mathcal{Z}) &= \|\mathbf{x}_{0}^{0}\|_{\mathbf{\Omega}_{0}}+\|\mathbf{m}_{0}^{0}\|_{\mathbf{\Omega}_{0}}{} \\
	& {}+ \sum_{s=0}^{S-1}(
	\sum_{i,j}^{}\|\mathbf{x}_{i}^{s}\ominus \mathbf{m}_{j}^{s'}\ominus \mathbf{z}_{i^{s}j^{s'}}^{mix}\|_{\mathbf{\Omega}_{i^{s}j^{s'}}^{mix}}^{2}{}\\
	& {}+ \sum_{i,j}^{}\|\mathbf{x}_{i}^{s}\ominus \mathbf{x}_{j}^{s'}\ominus \mathbf{z}_{i^{s}j^{s'}}^{odom}\|_{\mathbf{\Omega}_{i^{s}j^{s'}}^{odom}}^{2}{} \\
	& {}+ \sum_{i,j}^{}\|\mathbf{x}_{i}^{s}\ominus \mathbf{x}_{j}^{s'}\ominus \mathbf{z}_{i^{s}j^{s'}}^{node}\|_{\mathbf{\Omega}_{i^{s}j^{s'}}^{node}}^{2}{} \\
	& {}+\sum_{i,j}^{}\|\mathbf{m}_{i}^{s}\ominus \mathbf{m}_{j}^{s'}\ominus \mathbf{z}_{i^{s}j^{s'}}^{map}\|_{\mathbf{\Omega}_{i^{s}j^{s'}}^{map}}^{2}),
	\end{split}
	\end{equation}
	
	\section{Pose Graph Refinement}
	In this section, we present the two main components of our system: 1) multi-session localization and 2) pose graph refinement.
	
	\subsection{Multi-session Localization}
\begin{figure}[h]
		\centering
		\includegraphics[scale=0.13]{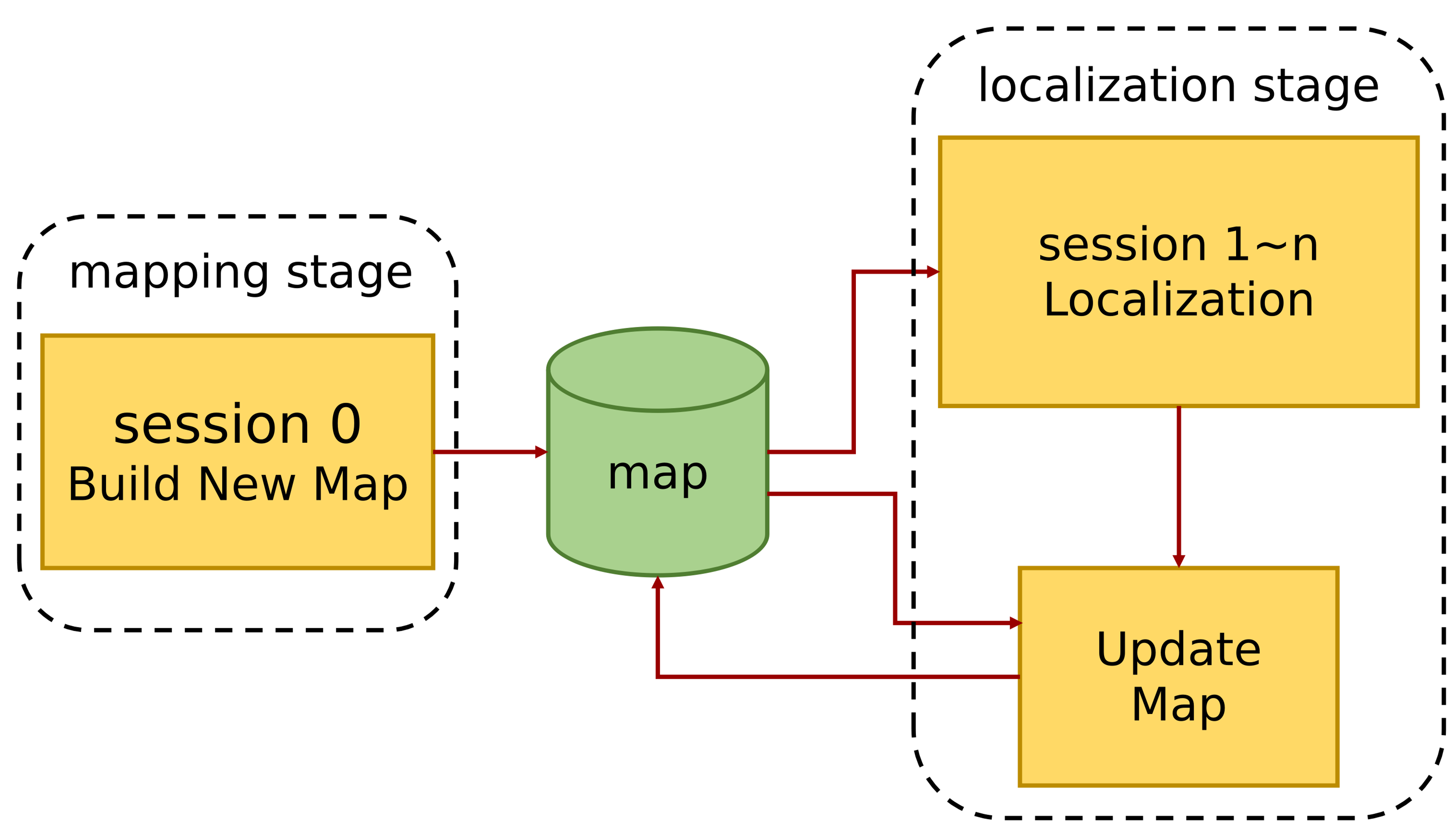}
		\caption{Schematic illustration of the map update process. A new map is firstly built during session 0 of the mapping stage. Given the pre-built map, robot then estimates its pose and updates the map in the following sessions of the localization stage. %Each localization session will repeat the same procedure we talked before for the long-term operations.
		}
		\label{fig:pipeline}
	\end{figure}

	The methodology of our map management procedure is based on a map update process as depicted in Fig. \ref{fig:pipeline}. A robot deployed into a new environment has to execute mapping (in session 0), collect the sensor data(including LiDAR, IMU and wheel encoder) and build the map representation of current environment. The map consists of several occupancy grid submaps, where each submap includes a fixed number LiDAR scans with corresponding poses, i.e. nodes. There are two advantages: 1) the single submap is immune to the global optimization because of local scan-to-submap matching; and 2) it is convenient to update the global map by trimming old submaps and adding new submaps to it. After the mapping stage, the robot performs localization task and creates new submaps from LLO. These submaps are always fresh and continuously recording the newest characteristics of current environment. Once a new submap is created, it is transmitted to the PGR for the subsequent map updating. Apart from LLO, the sensor data are input to GLM. GLM is responsible for calculating the relative measurement constraints between scans and submaps in global map and outputting the constraints to PGR. PGR is the key subsystem of our proposed framework, it receives the fresh submaps and constraints from LLO and GLM, respectively. PGR consists of three modules: submaps trimming, pose graph sparsification and pose graph optimization. It maintains the newest submaps by replacing the stale submaps in the old session. Furthermore, in order to keep the sparse characteristic of the pose graph, the relevant stale submaps, nodes and constraints are removed. The remaining submaps from PGR are transmitted to the global submap database for the subsequent localization task. We call this process "Map Updating".
	
	Each localization session repeats the aforementioned procedure to estimate the pose of the robot and achieve the updated map.

	\subsection{Pose Graph Refinement}
	\textit{1) Submaps Trimming:} In the context of lifelong localization, new submaps are added to the global map instead of stale submaps whenever the robot is re-entering previously visited terrain. The key idea is to trim the old submaps to limit its number. Most of the existing approaches rely on the environment change detection \cite{b10}, \cite{b11}. They need to find out when it is time to update the localization map by comparing the old and latest maps cell-by-cell. To mitigate computation complexity, we adopt the method from \cite{b2} that computes the overlapping ratio of the stale submap. In case the ratio is below a defined threshold, the old submaps are not deleted. Otherwise, they are marked as trimmed and deleted in the following pose graph sparsification module. The fresh submaps are added to the pose graph regardless of the status of the old submap. The advantage of this method is that we can keep a constant computational time for a robot working in a fixed area.
	
	\textit{2) Pose Graph Sparsification:} 
	%The texture of the submap and its corresponding pose node must be deleted from memory in the submap trimming module.
	%The previous subsection presented the method for finding the submaps to trim. Not only the texture of map should be deleted from memory, but also its pose node must be removed from pose graph. 
	A direct way of discarding a trimmed submap is to throw all the constraints and nodes that are connected with the submap. However, such method will lose a lot of information(e.g., the relative measurements between submap and nodes) about the pose graph, and thus resulting in the instability of the graph. Marginalization is a effective way to mitigate this problem. To avoid introducing new edges between all pairs of variables (dense fill-in), which reduces the sparsity of the graph and greatly increase the computational complexity, we adopt the Chow-Liu tree \cite{b13} to approximate the individual elimination cliques as sparse tree structures. 
	
\begin{figure*}[]
		\centering
		\includegraphics[scale=0.5]{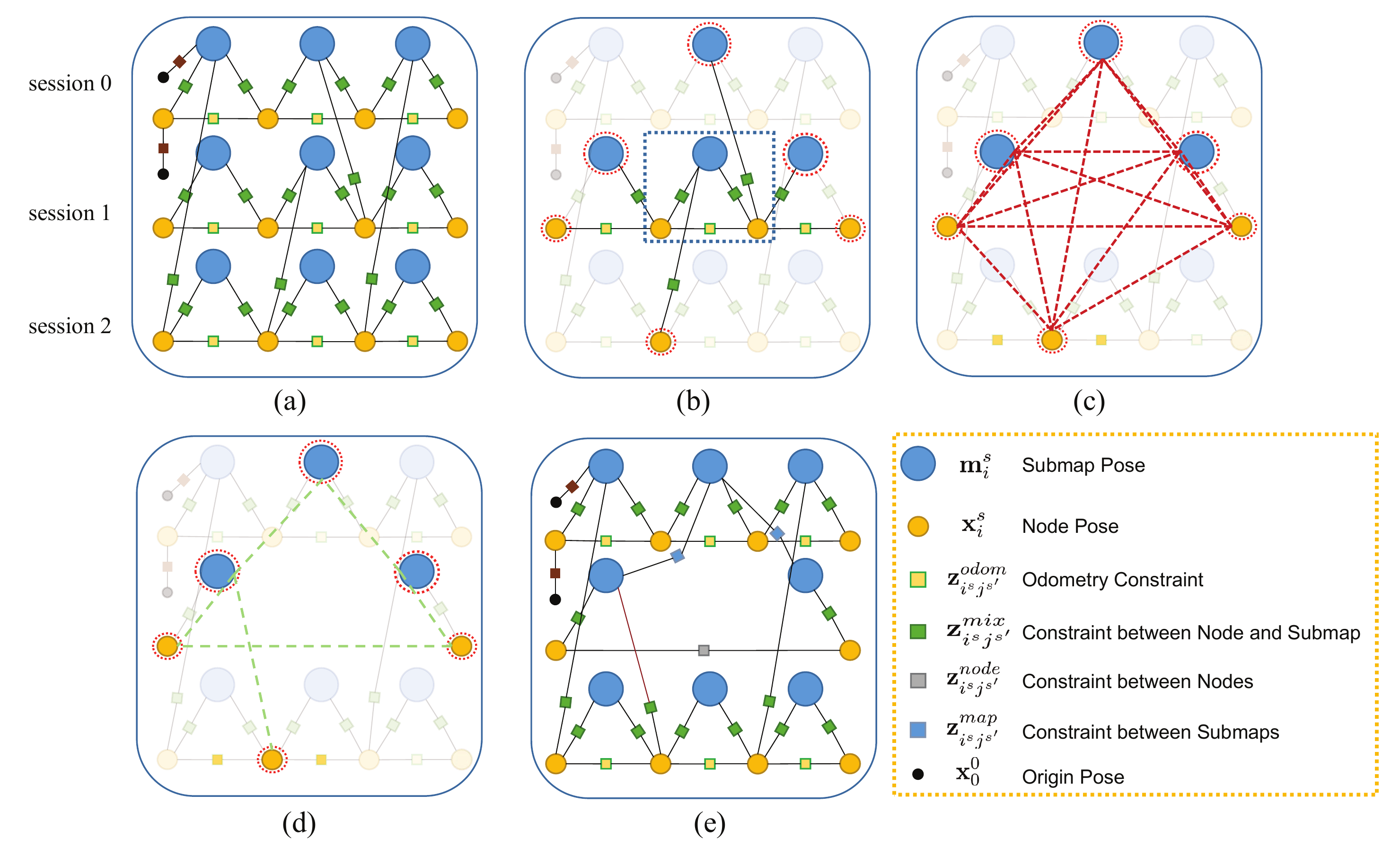}
		\caption{An example to illustrate the graph sparsification based on the Chow–Liu maximum-mutual-information spanning tree. (a): the original pose graph without submap removal. (b): the central submap $m_{1}^{1}$ is selected to be trimmed along with its nodes: $x_{1}^{1}$ and $x_{2}^{1}$(blue dotted rectangle). The red dotted circles highlight the relevant submaps and nodes that have constraints with blue rectangle. (c): Belief after having marginalized out $m_{1}^{1}$, $x_{1}^{1}$ and $x_{2}^{1}$. The former neighbors of them form an elimination clique making the graph dense. (d): Belief resulting from Chow-Liu tree approximation of the elimination clique. (e): Merge the elimination clique into original pose graph.}
		\label{fig:graph}
	\end{figure*}

	Fig. \ref{fig:graph} illustrates the sparsification process. Given an original pose graph (Fig. \ref{fig:graph}(a)), a submap with its two nodes are slated for removal in Fig. \ref{fig:graph}(b) (blue dotted rectangle). We extract the related submaps and nodes (dots with red dotted circles in Fig. \ref{fig:graph}(b)) as a local factor graph. After marginalizing out the submap and nodes, the former neighbors form an elimination clique making the graph dense (Fig. \ref{fig:graph}(c)). By locally approximating the density $p(\mathbf{x}_{1},...,\mathbf{x}_{n})$ with a probability distribution $q(\mathbf{x}_{1},...,\mathbf{x}_{n})$ such that each variable is only conditioned upon one of the other variables, we can obtain the sparse local factor graph:
	\begin{equation}
	\begin{split}
	p(\mathbf{x}_{1},...,\mathbf{x}_{n})&=p(\mathbf{x}_{1})\prod_{i=2}^{n}p(\mathbf{x}_{i}|\mathbf{x}_{i-1},...,\mathbf{x}_{1})\\
	&\approx p(\mathbf{x}_{1})\prod_{i=2}^{n}p(\mathbf{x}_{i}|\mathbf{x}_{i-1})\\
	&= q(\mathbf{x}_{1},...,\mathbf{x}_{n}),
	\end{split}
	\end{equation}
	where $\mathbf{x}_{1}$ is the root variable of the CLT, and $\mathbf{x}_{i-1}$ is the parent of $\mathbf{x}_{i}$. The pairwise conditional distributions are selected such that the Kullback-Leibler divergence (KLD) between the original distribution and the CLT approximation is minimized (Fig. \ref{fig:graph}(d)). It is given by:
		\begin{equation}
		D_{KL}(p||q)=\int_{\mathbf{x}}p(\mathbf{x})\mathrm{log}\frac{p(\mathbf{x})}{q(\mathbf{x})}d\mathbf{x}.
		\end{equation}
	
	Lastly, we merge the approximate clique into the original graph as depicted in Fig. \ref{fig:graph}(e). The above procedures are described in Algorithm 1. It should be noted that GLM only produces the node-to-submap constraints and inputs them into PGR (see the original pose graph in Fig. \ref{fig:graph}(a)). However, the node-to-node constraints and submap-to-submap constraints are induced in sparse pose graph (Fig. \ref{fig:graph}(e)). This is because all nodes and submaps are treated as the same variable type in the elimination clique. By merging the clique into original graph, the variables and pairwise constraints are converted into the corresponding nodes, submaps and their constraints. 
	
	After graph sparsification, we execute optimization (see the cost function in Eq. (1)) by using the Google Ceres Solver \cite{b19}.
	
			\begin{algorithm}[hb]
\SetKwData{Left}{left}\SetKwData{This}{this}\SetKwData{Up}{up}
\SetKwFunction{Concatenate}{Concatenate}\SetKwFunction{Kmeans}{K-means}
\SetKwInOut{Input}{input}\SetKwInOut{Output}{output}
\Input{\small{Submaps $\{\mathbf{m}_{0},...,\mathbf{m}_{k}\}$ and pose graph $\mathcal{G}$}}
\Output{\small{Trimmed submaps $\{\mathbf{m}_{0},...,\mathbf{m}_{k}\}$ and sparsified pose graph $\mathcal{G}$}}
\ForAll{\small{$\mathbf{m}_{i} \in \{\mathbf{m}_{0},...,\mathbf{m}_{k}\}$}}
{
\small{Get nodes $\mathcal{X}_{intra} = \{\mathbf{x}_{0},...,\mathbf{x}_{m}\}$ belonging to $\mathbf{m}_{i}$} \;
\small{Find all constraints $\mathcal{Z}$, nodes $\mathcal{X}$ and submaps $\mathcal{M}$ related to $\mathbf{m}_{i}$} \;
\small{Construct an iSAM graph and add $\mathbf{m}_{i}$, $\mathcal{X}_{intra}$, $\mathcal{X}$, $\mathcal{M}$ as nodes and $\mathcal{Z}$ as factors to the graph} \;
\tcp{\small{see Eq. (2) and (3)}}
\small{RemoveNodeBasedCLT$(\mathbf{m}_{i})$} \;
\ForAll{\small{$\mathbf{x}_{i}$ in $\mathcal{X}_{intra}$}}
{
\small{RemoveNodeBasedCLT$(\mathbf{x}_{i})$}
}
\small{delete $\mathcal{X}_{intra}$, $\mathbf{m}_{i}$ from $\mathcal{G}$ and merge the remaining factors from the iSAM graph to $\mathcal{G}$} \;
}
\caption{\small{Submap trimming and pose graph sparsification}}
\end{algorithm}
	
	\section{Dataset And Experiment}
	
	\subsection{Experimental Setup and Dataset}
\begin{figure}[htpb]
		\centering
		\includegraphics[scale=0.3]{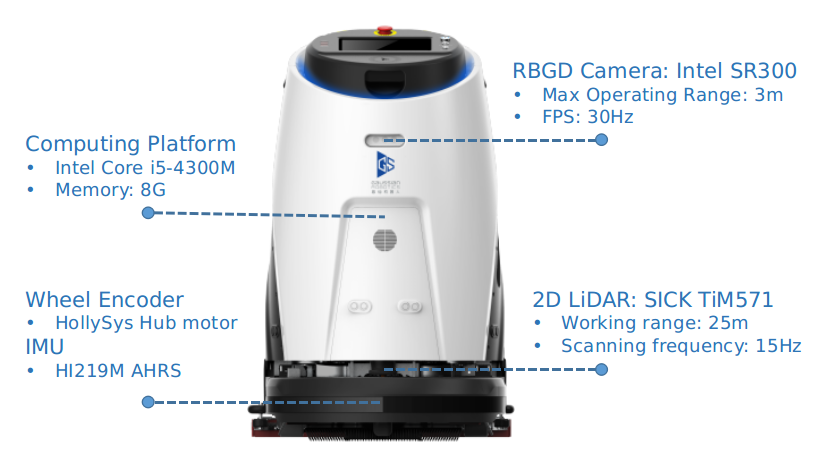}
		\caption{Our cleaning robot is equipped with a SICK TiM571 2D LiDAR. It is equipped with a magnetic encoders and a HI219M AHRS (with on-board 3-aixs accelerometer, 3-aixs gyroscope and 3-aixs compass) for dead-reckoning. An Intel SR300 RGBD camera is used for perception and obstacle avoidance. Our algorithm is performed on Intel Core i5-4300M 2 cores and 4 threads.}
		\label{fig:sensor}
	\end{figure}

	To validate our algorithm, we deploy a three-wheeled cleaning robot in the supermarket as shown in Fig. \ref{fig:sensor}. The algorithm modules including localization, navigation and perception are performed on an industrial computer with Intel i5-4300M CPU and 8G memory. Over a period of a month the robot executed cleaning tasks in the indoor environment with random start position.
	We choose a supermarket located in Beijing, China as our experimental site because of its highly dynamic flow of people, moving carts, goods, etc. That is a huge challenge for accuracy and stability of any localization algorithm. 
	
	Because the test data including scene changes in real world are so valuable for designing more robust localization system, we collected them as dataset and released it in order to accelerate lifelong SLAM research in the community. The scenes in the dataset include markets, garages and offices, recored by 2D LiDAR(Sick TiM571) and 3D LiDAR(RoboSense RS-LiDAR-16). The dataset also records other sensor data, including wheel encoder and IMU. For each data sequence, we carefully calibrated the intra-device extrinsics, especially the transform between LiDAR and odometer. We also provided the ground truth poses in a rate of 10 Hz, which are manually verified by comparing the LiDAR scans with the pre-built consistent map.
	
	\subsection{Algorithm Evaluation}
		The initial map consists of 550 submaps in session 0, covering an area of over 10000 square meters. These submaps are cached in memory for further localization and cannot be trimmed. To balance the accuracy and computation load, we executed graph optimization every 5 times of sparsification and restricted maximum number of iterations in Ceres Solver to 50.
	\begin{figure*}[]
	    \includegraphics[scale=0.64]{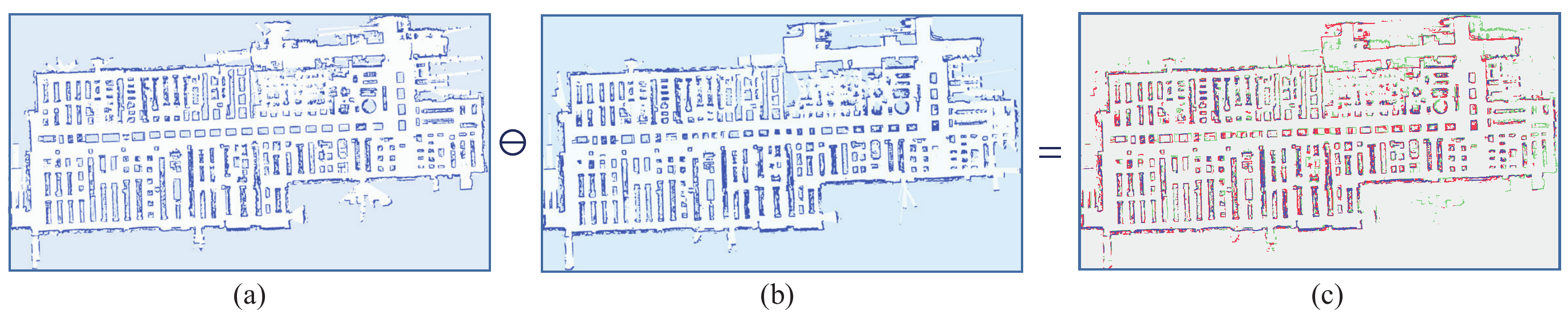}
	    \caption{Results of map change after one month running in market. (a): The pre-built map. (b): The updated map. (c): Comparison(define $\ominus$ to execute such operation) of (a) and (b). The blue points are obstacles that remain unchanged. The light green points are obstacles which have been removed and red points are newly added obstacles.}
	\label{fig:map}
	\end{figure*}
	
	\begin{figure*}[]
		\centering
		\includegraphics[scale=0.82]{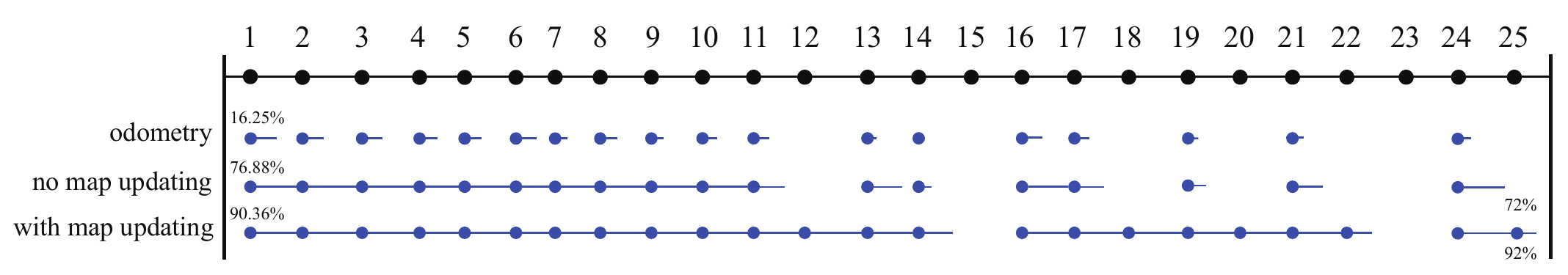}
		%	\centerline{\includegraphics{evoluate.eps}}
		\caption{Localization testing results with 25 sets data from market. The percentage value on the top left of each algorithm is MRCL and on the bottom right is CRI.}
		\label{fig:evaluate}
	\end{figure*}
	
\begin{figure*}[ht]
		\centering
		\includegraphics[scale=0.5]{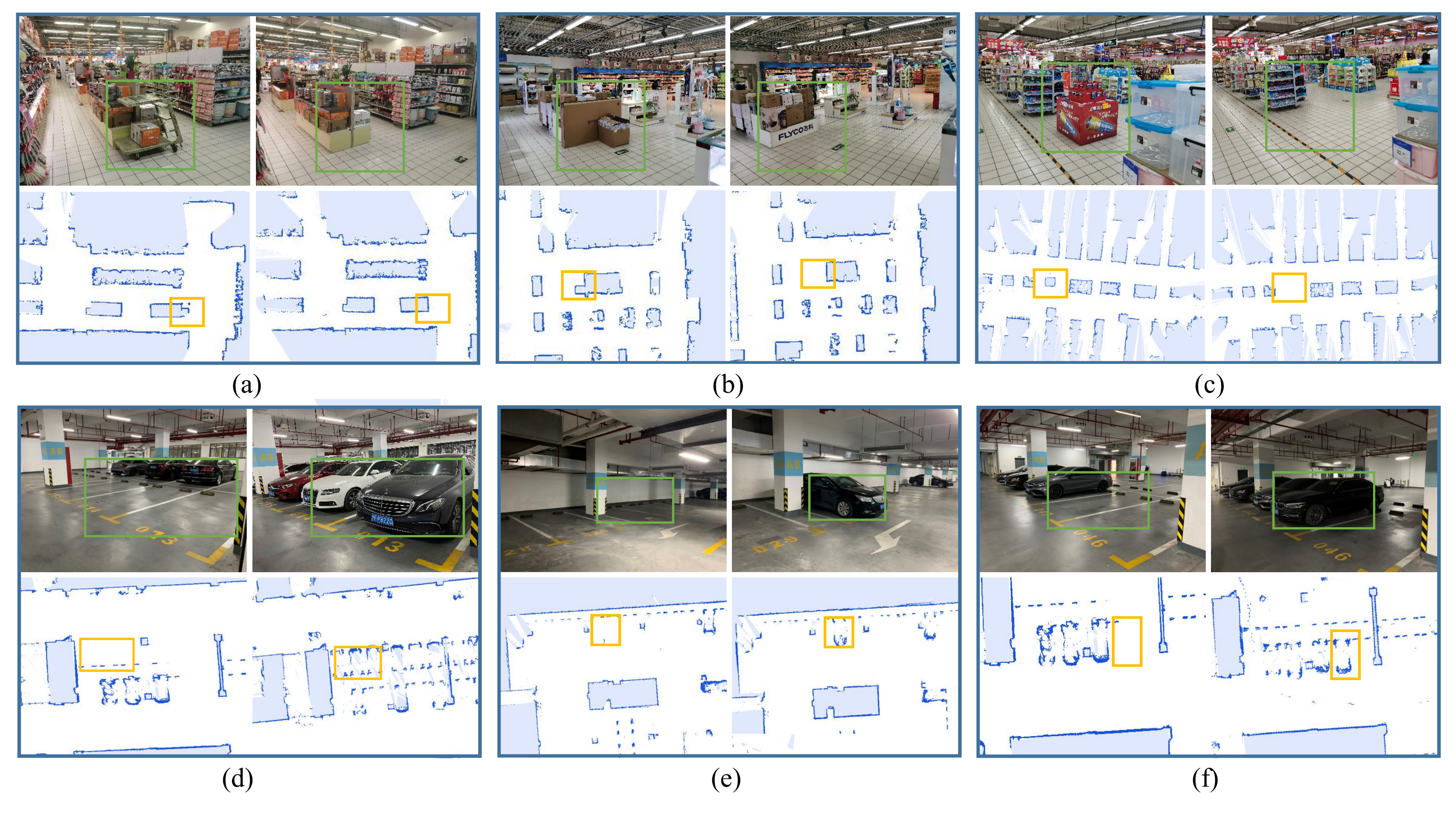}
		\caption{Examples of environment change and corresponding map updating experiment. The result in (a), (b) and (c) are collected from market. And (d), (e) and (f) are from garage.}
		\label{fig:garage}
	\end{figure*}

	\begin{itemize}
		\item \textbf{Environmental Change} Based on our map updating method, the robot collects new submaps and trims the old ones during the localization stage. The fresh submaps update the latest representation of current environment
		whenever the scene changes. Fig. \ref{fig:garage} shows the experimental results. (a)(b)(c) are collected from market. The upper left image and upper right image in each column show approximately the same place captured at different time. The lower left and right images show the corresponding map updating results. It should be noted that the appearance of map is not the same as the submaps because we concatenated the submap slices into single occupancy grid map for convenient view. In addition, we test our algorithm in garage scenes and (e)(f)(g) shows the results.
		
		Fig. \ref{fig:map} shows the results of map change after running for one month in the market. (a) is pre-built map in mapping stage and (b) is the updated map based on (a). We compare the differences between these two maps as shown in (c). The blue points are the unchanging obstacles in all of time. The light green points are the obstacles that have been removed from the scene, and red points are new added obstacles. We also define the map change rate (MCR) to evaluate the degree of environment change:
		\begin{equation}
			MCR = \frac{(m_{r}+m_{g})*0.5}{(m_{r}+m_{g})*0.5+m_{b}},
		\end{equation}
		where $m_{r}$, $m_{g}$, $m_{b}$ is the number of red, green and blue pixel respectively. The MCR of our map is 51.62\%, meaning that more than half of the features have been changed in this scene. This metric can reveal the environment change easily.
		
		\item \textbf{Localization Performance Evaluation} We execute the robot cleaning task in the market and collect the data sequence separately for comparison of our algorithm with the general SLAM algorithm \cite{b2}. For real scene, we cannot obtain the ground-truth poses. Therefore, we design separate metrics to evaluate the correctness and accuracy, respectively.
		
     	\textit{Correctness Rate of Initialization(CRI)}. Prior to executing a localization task, the robot should receive the initial pose estimation and match the map with collected LiDAR sensor data. A better initialization result can be obtained from a newer map. We define a index of correct initialization as:
     	\begin{equation}
			CRI=\frac{\sum_{k=0}^{N-1}c_{k}}{N},
     	\end{equation}
     	where $c_{k}$ can be defined as:
     	\begin{equation}
        c_{k}=
        \begin{cases}
        1, & \text{if\ initialization\ succeed} \\
        0, & \text{otherwise}
        \end{cases}
        \end{equation}

     	\begin{table*}[]
			\setlength{\tabcolsep}{0.84mm}{
				\caption{AVERAGE MATCHING SCORE COMPARISON}
				\begin{tabular}{@{}clllllllllllcllcllclclccll@{}}
					\toprule
					\multicolumn{1}{l}{Data Sequence} & \multicolumn{1}{c}{1} & \multicolumn{1}{c}{2} & \multicolumn{1}{c}{3} & \multicolumn{1}{c}{4} & \multicolumn{1}{c}{5} & \multicolumn{1}{c}{6} & \multicolumn{1}{c}{7} & \multicolumn{1}{c}{8} & \multicolumn{1}{c}{9} & \multicolumn{1}{c}{10} & \multicolumn{1}{c}{11} & 12                                & \multicolumn{1}{c}{13} & \multicolumn{1}{c}{14} & 15         & \multicolumn{1}{c}{16} & \multicolumn{1}{c}{17} & 18                                & \multicolumn{1}{c}{19} & 20                                & \multicolumn{1}{c}{21} & 22                                & 23 & \multicolumn{1}{c}{24} & 25                    \\ \midrule
					{[}2{]}                          & 0.73                  & \textbf{0.72}         & 0.76                  & \textbf{0.76}         & 0.72                  & 0.72                  & 0.68                  & 0.67                  & 0.71                  & 0.62                   & 0.52                   & \textit{-}                        & 0.68                   & 0.18                   & -          & \textbf{0.70}          & 0.43                   & -                                 & 0.16                   & -                                 & 0.44                   & -                                 & -  & 0.42                   & \multicolumn{1}{c}{-} \\
					Ours                              & \textbf{0.74}         & 0.70                  & \textbf{0.78}         & 0.75                  & \textbf{0.74}         & \textbf{0.74}         & \textbf{0.73}         & \textbf{0.72}         & \textbf{0.72}         & \textbf{0.72}          & \textbf{0.69}          & \multicolumn{1}{l}{\textbf{0.68}} & \textbf{0.77}          & \textbf{0.48}          & \textbf{-} & 0.69                   & \textbf{0.59}          & \multicolumn{1}{l}{\textbf{0.74}} & \textbf{0.68}          & \multicolumn{1}{l}{\textbf{0.68}} & \textbf{0.72}          & \multicolumn{1}{l}{\textbf{0.26}} & -  & \textbf{0.72}          & \textbf{0.58}         \\ \bottomrule
			\end{tabular}}
		\label{table_1}
		\end{table*}
     	
     	\textit{Mileage Rate of Correct Localization(MRCL)}. By comparing the robot poses output and pre-built map, we can manually verify the correctness of the robot localization trajectory. We use MRCL as the metric. For a sequence from $t_{0}$ to $t_{max}$, we define:
     	\begin{equation}
			MRCL=\frac{correct\_mileage(t_{0},t_{max})}{total\_mileage(t_{0},t_{max})},
     	\end{equation}
     	where $correct\_mileage()$ is used to calculate the mileage of correct trajectory and $total\_mileage()$ is used to calculate the total mileage.
     	
     	\textit{Average Matching Score(AMS)}. Our method of PGR module is the same with ``Global SLAM'' module in \cite{b2}, where building constraints between nodes and submaps relied on a ``branch and bound (BNB)'' mechanism to work at different grid resolutions. Therefore, we use the BNB score to define whether a node is correctly matched against a submap. Usually, we set a minimum score(in our experiment, the score is 0.5) to accept a good enough proposal and fed it into PGR. Otherwise, we discard this constraint candidate. Formally, we define a average score of matching as:
		\begin{equation}
			AMS=\frac{\sum_{i=0}^{M-1}s_{i}}{M},
		\end{equation}
		where $M$ is the number of matching. $s_{i}$ is the i-th BNB score.

		The results are visualized in Fig. \ref{fig:evaluate}, with blue line segments indicating the correct localization trajectory and blue dots indicating whether the initialization is successful or not. Each black dot on the top line represents the start of a data sequence. The blue line segments start from the blue dots means robot initializes successfully and runs for a while. The length of blue line segments is proportional to a correct mileage. We tested 25 data sequences with three algorithms: original odometry output, method from \cite{b2} and ours. The CRI and MRCL from Fig. \ref{fig:evaluate} demonstrate that our method is more robust enough for localization in the challenging environment compared with \cite{b2}. However, it should be noted that, in sequence 15 and 23, our method failed to do initialization. This is because the environment is full of pedestrians and the score of scan-matching does not exceed initialization BNB score. The AMS comparison result is listed in Table I. The results show that with longtime map updating, our method achieves higher matching score for robust constraints building based on fresh map.
		
    	We display the trajectory comparison from 17th data sequence as shown in Fig. \ref{fig:traj}. The blue line plots the fusion result of wheel encoder and IMU. The green line is from our method and the red line is the result of \cite{b2}. It demonstrates that the trajectory of ours is smoother than \cite{b2}. The reason is that, with map updating and higher matching score, the valid constraint measurements are much more  dense than \cite{b2}, which only computes constraints based on the stale map.
    	
    \begin{figure}[htpb]
		\centering
		\includegraphics[scale=0.5]{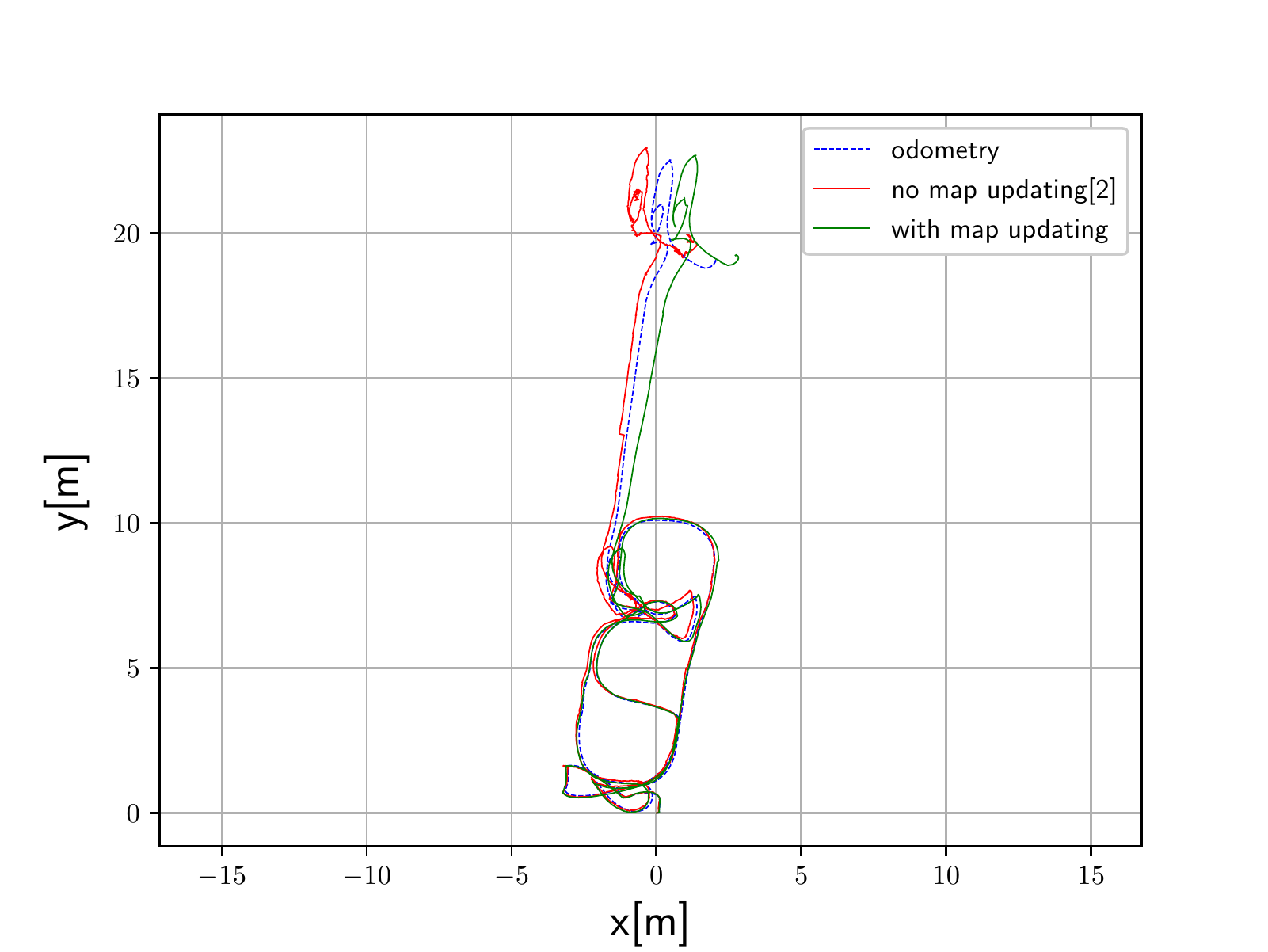}
		%	\centerline{\includegraphics{traj_compare.eps}}
		\caption{Comparison of trajectories from 17th data sequence.}
		\label{fig:traj}
	\end{figure}
	
	\begin{figure}[]
		\centering
		\includegraphics[scale=0.43]{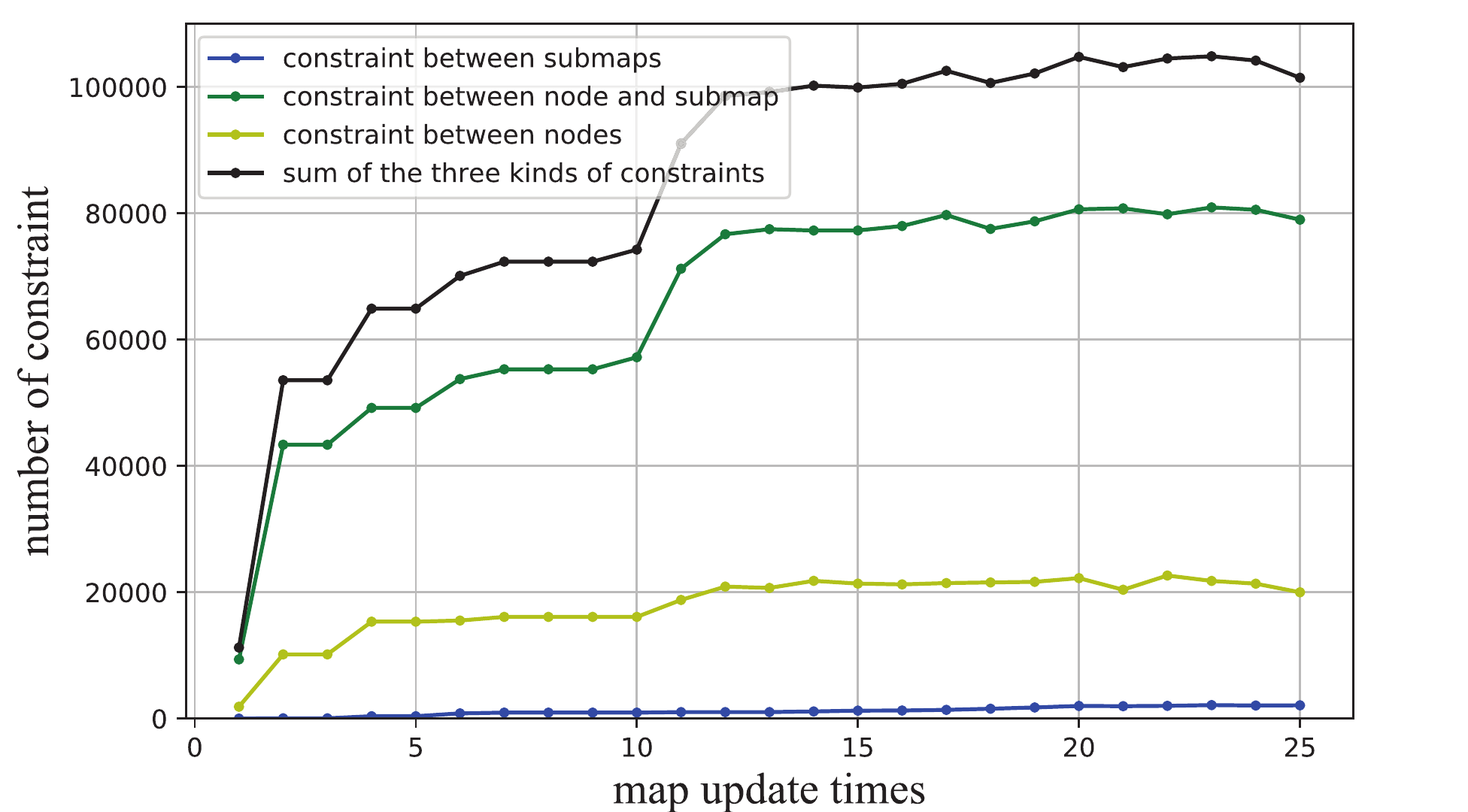}
		%\centerline{\includegraphics{contraint_crop_5.eps}}
		\caption{Lifelong SLAM graph complexity. The number of constraints remains roughly constant after 15 times of map updating.}
		\label{fig:cons}
	\end{figure}

		\item \textbf{Computational Complexity Evaluation} The size of the pose graph has a direct influence to the runtime and the memory complexity of the SLAM system and typically grows over time. Therefore, it is desirable to prevent the scale of the graph from growing unbounded. To check the graph scale, we record the essential elements of the pose graph in each map updating. We see in Fig. \ref{fig:node} that the PGR module limit the number of nodes and submaps to less than 22000 nodes and 650 submaps. As shown in Fig. \ref{fig:cons}, the total number of constraints is under 110000. Fig. \ref{fig:usage} depicts the average CPU utilization and memory usage of our SLAM algorithm while robot executed localization task. The computation load is positively associated with the scale of graph. It means that our method will not occupy unlimited resource of the system. 

	\end{itemize}

	\begin{figure}[]
		\includegraphics[scale=0.43]{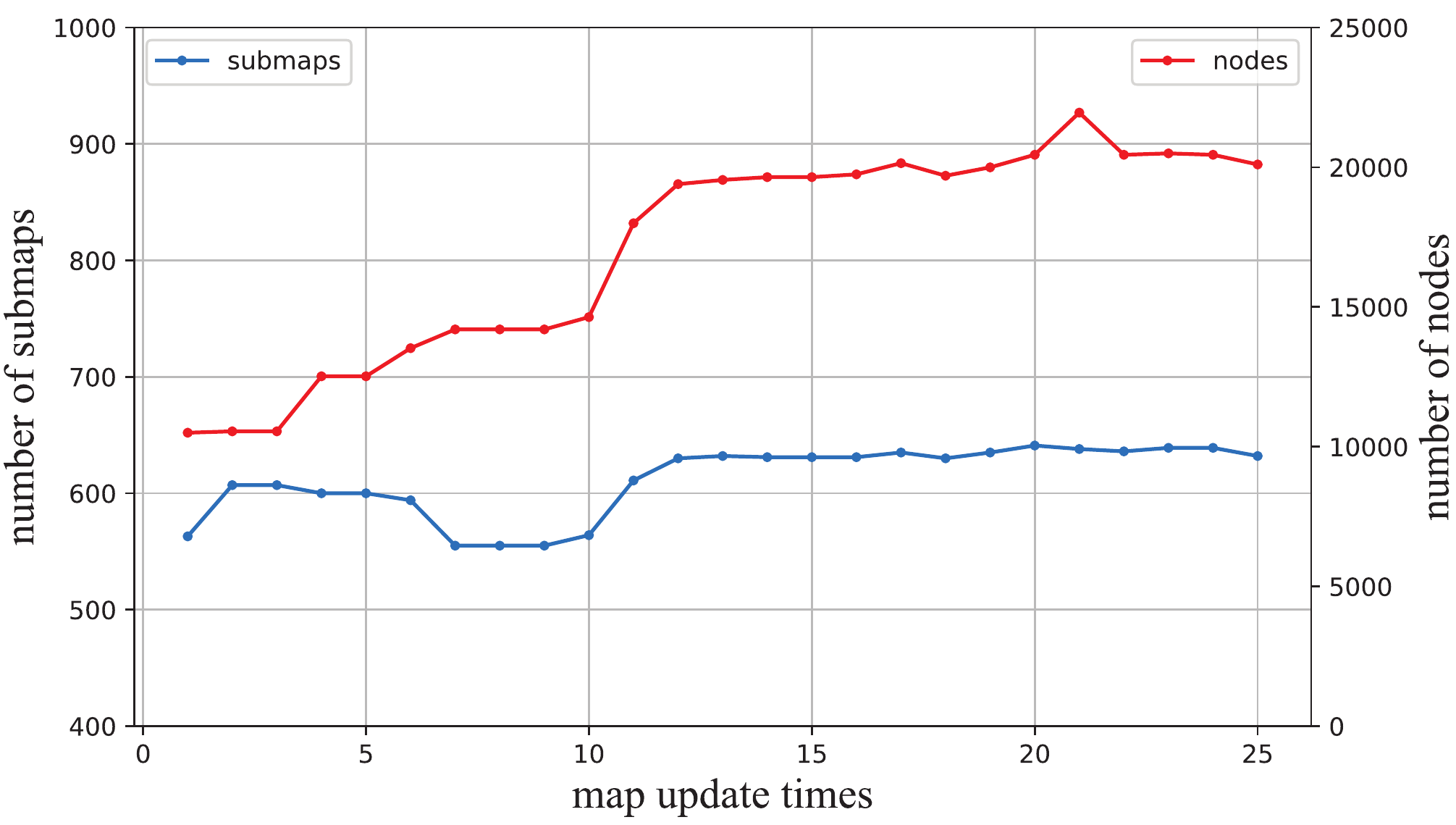}
		%\centerline{\includegraphics{node_crop_7.eps}}
		\caption{Lifelong SLAM graph complexity. The number of nodes and submaps remains roughly constant after 15 times of map updating.}
		\label{fig:node}
	\end{figure}

	\begin{figure}[htpb]
		\centering
		\includegraphics[scale=0.55]{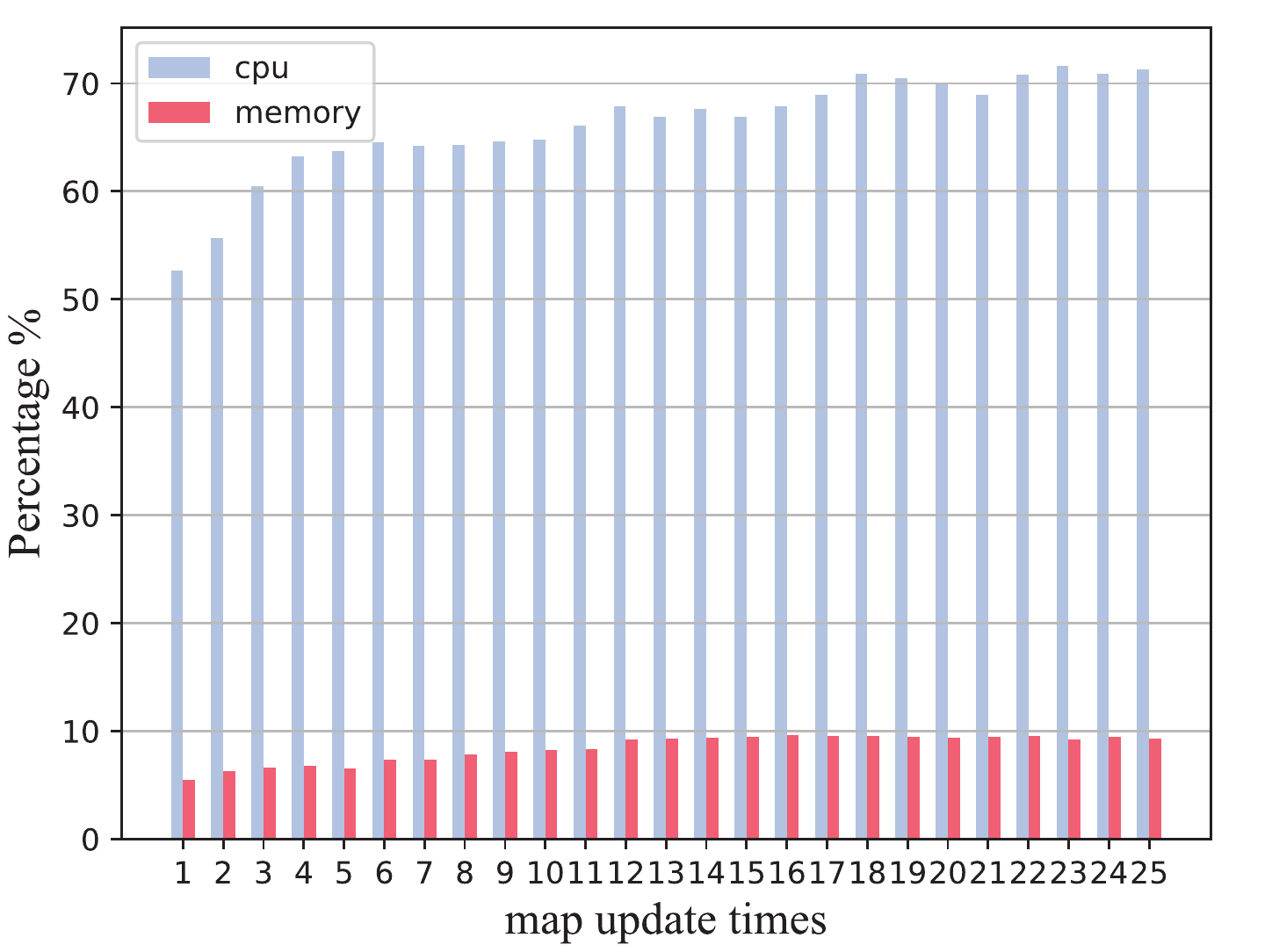}
		%\centerline{\includegraphics{usage.eps}}
		\caption{Computation load changes with 25 times of map updating.}
		\label{fig:usage}
	\end{figure}
	
	\section{Conclusions And Future Work}
	To address the problem of environment change in real-world, we present a complete lifelong SLAM framework. The proposed method exploits the multiple localization sessions and map updating strategy, which can be used to track the scene change and achieve an up-to-date map. We also propose a submap-based graph sparsification method based on Chow-Liu maximum-mutual-information spanning tree to balance computational complexity and localization accuracy. We comprehensively validate our method in real supermarket for more than a month. The experiment demonstrates that our method is quite valuable to be deployed in real world. Besides, we release our lifelong SLAM dataset to accelerate robust SLAM research in these scenes.
	
	However, our approach does not at present take into account the unexpected drastic change of the environment. The change may cause localization drift because there are no valid constraint measure between incoming scans and map. In the future, we plan to explore more robust localization algorithm to conquer this problem.
	
	\section*{Acknowledgement}
    Authors appreciate the great help from our outstanding colleagues, Wenjing Chen, Pengfei Du, Shiwei Wang, Linbing Zen, Bo Wu, Haoxuan Tan, Guolin Li, Aiwei Yin and Youji Zhu.
    
    The video demo is available at   \url{https://youtu.be/U7b0zc_jK74}.
	
\end{document}